\documentclass{article}

\usepackage[preprint]{neurips_2024}

\usepackage[table]{xcolor}
\usepackage[utf8]{inputenc} 
\usepackage[T1]{fontenc}    
\usepackage{hyperref}       
\usepackage{url}            
\usepackage{booktabs}       
\usepackage{amsfonts}       
\usepackage{nicefrac}       
\usepackage{microtype}      
\usepackage{xcolor}         
\usepackage{caption}      
\usepackage{hyperref}     
\usepackage{lipsum}       

\usepackage{graphicx}
\usepackage{float}
\usepackage{xspace}
\usepackage{amsmath} 
\usepackage{cleveref}
\usepackage{natbib}
\setcitestyle{numbers,square}
\usepackage{wrapfig}

\usepackage{multirow}
\definecolor{myblue}{RGB}{210, 225, 255}
\definecolor{logo}{HTML}{00a6fb}

\title{\includegraphics[height=0.8em]{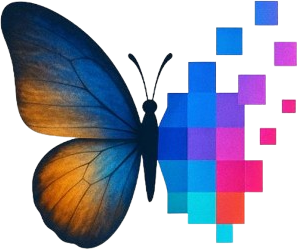} \textcolor{logo}{Echo-4o}: Harnessing the Power of GPT-4o Synthetic Images for Improved Image Generation}

\author{
\textbf{Junyan Ye}\textsuperscript{1,2 *},
\textbf{Dongzhi Jiang}\textsuperscript{3*},
\textbf{Zihao Wang}\textsuperscript{2},
\textbf{Leqi Zhu}\textsuperscript{1},
\textbf{Zhenghao Hu}\textsuperscript{2},
\textbf{Zilong Huang}\textsuperscript{2},
\\
\textbf{Jun He}\textsuperscript{2},
\textbf{Zhiyuan Yan}\textsuperscript{4},
\textbf{Jinghua Yu}\textsuperscript{2},
\textbf{Hongsheng Li}\textsuperscript{3 \dag},
\textbf{Conghui He}\textsuperscript{1 \dag},
\textbf{Weijia Li}\textsuperscript{2 \dag}
\\
{\normalsize $^1$Shanghai Artificial Intelligence Laboratory \quad $^2$Sun Yat-sen University} 
\\
{\normalsize $^3$CUHK MMLab \quad $^4$Peking University} 
\\ 
\vspace{2mm}
\parbox{\textwidth}{
\centering
\begin{tabular}{ll}
\raisebox{-0.15em}{\includegraphics[height=1.05em]{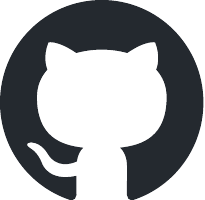}} \textbf{Github:} & \url{https://github.com/yejy53/Echo-4o} \\
\raisebox{-0.15em}{\includegraphics[height=1.05em]{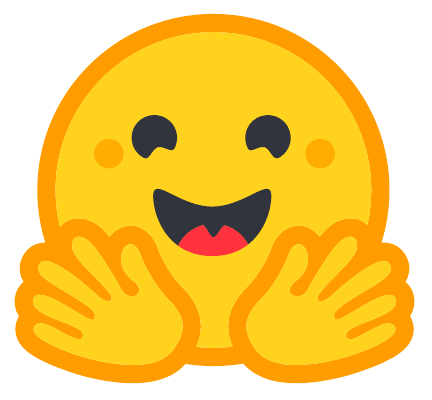}} \textbf{Dataset:} & \url{https://huggingface.co/datasets/Yejy53/Echo-4o-Image/} \\
\raisebox{-0.15em}{\includegraphics[height=1.05em]{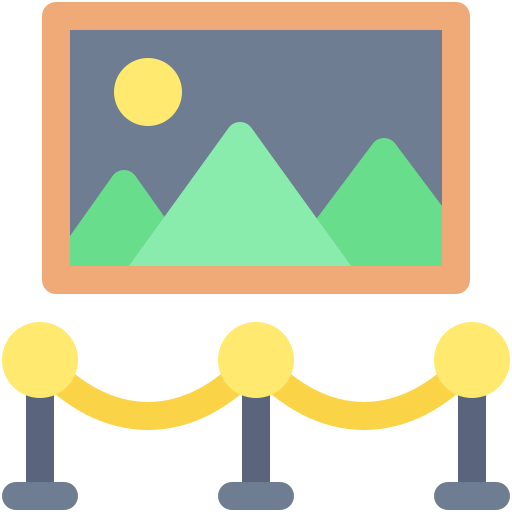}} \textbf{Gallery:} & \url{https://yejy53.github.io/Echo-4o} \\
\end{tabular}
}
}

\begin{document}

\maketitle

\begin{abstract}
Recently, GPT-4o has garnered significant attention for its strong performance in image generation, yet open-source models still lag behind. Several studies have explored distilling image data from GPT-4o to enhance open-source models, achieving notable progress. 
However, a key question remains: given that real-world image datasets already constitute a natural source of high-quality data, why should we use GPT-4o-generated synthetic data?
In this work, we identify two key advantages of synthetic images. First, they can complement rare scenarios in real-world datasets, such as surreal fantasy or multi-reference image generation, which frequently occur in user queries. Second, they provide clean and controllable supervision. Real-world data often contains complex background noise and inherent misalignment between text descriptions and image content, whereas synthetic images offer pure backgrounds and long-tailed supervision signals, facilitating more accurate text-to-image alignment. Building on these insights, we introduce Echo-4o-Image, a 180K-scale synthetic dataset generated by GPT-4o, harnessing the power of synthetic image data to address blind spots in real-world coverage. Using this dataset, we fine-tune the unified multimodal generation baseline Bagel to obtain Echo-4o. In addition, we propose two new evaluation benchmarks for a more accurate and challenging assessment of image generation capabilities: GenEval++, which increases instruction complexity to mitigate score saturation, and Imagine-Bench, which focuses on evaluating both the understanding and generation of imaginative content. Echo-4o demonstrates strong performance across standard benchmarks such as GenEval, DPG, and OmniContext, as well as our proposed benchmarks. Moreover, applying Echo-4o-Image to other foundation models (e.g., OmniGen2, BLIP3-o) yields consistent performance gains across multiple metrics, highlighting the dataset’s strong transferability.

\end{abstract}

\begin{figure}[ht]
  \centering
  \includegraphics[width=\linewidth]{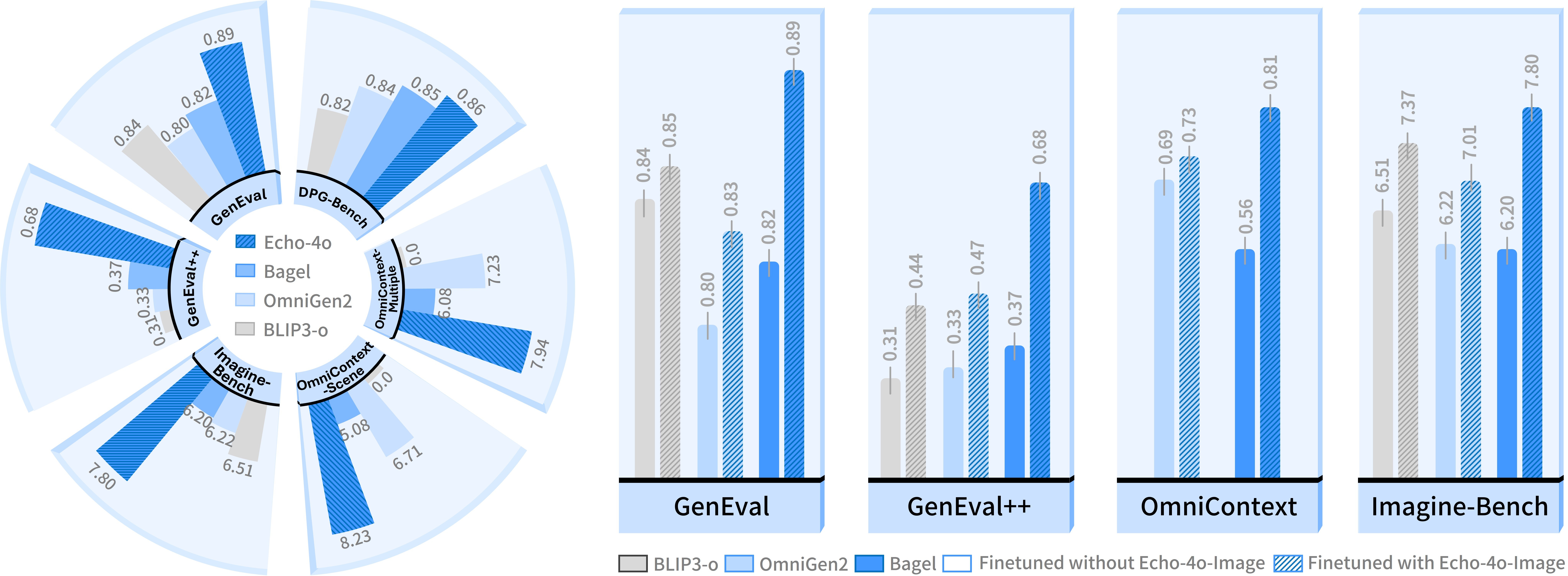}
  \caption{Performance gains of Echo-4o and transferability of Echo-4o-Image across models.}
  \label{fig:teaser}
\end{figure}

\section{Introduction}

Recently, unified multimodal generative models like GPT-4o~\cite{gpt4o} have attracted increasing attention, not only for their strong performance on the common generation tasks such as text-to-image generation and image editing, but also for its extraordinary understanding capability to perform free-form image manipulation~\cite{yan2025gpt}. 
Following this trend, a series of open-source unified models, including BLIP3-o~\cite{chen2025blip3}, Bagel~\cite{bagel}, and OmniGen2~\cite{wu2025omnigen2}, have been introduced. These models also showcase advanced proficiency in diverse multimodal generation tasks, highlighting the promising future of this research direction. Despite recent progress, current open-source unified models still exhibit a significant gap in overall generation quality compared to GPT-4o~\cite{gpt4o}, particularly in terms of instruction alignment, imaginative generation, and multi-reference image composition.

Similar challenges have been observed in the text or image understanding tasks of large language models (LLMs)~\cite{OpenAI2023ChatGPT, touvron2023llama, dubey2024llama3herdmodels, bai2023qwentechnicalreport} and multimodal large language models (MLLMs)~\cite{openai2023gpt4v, llava, bai2025qwen25vl}. A promising approach to mitigate this gap involves leveraging stronger models to synthesize high-quality data for smaller models~\cite{vicuna2023,llava,chen2023sharegpt4v,chen2024allava}. For instance, DeepSeek-R1's~\cite{guo2025deepseek} reasoning trajectories have proven highly effective in transferring advanced reasoning capabilities to smaller models, while ShareGPT-4V~\cite{chen2023sharegpt4v} utilizes GPT-4V to generate high-quality vision-language pairs. This strategy has recently been extended to the image generation domain. For example, BLIP3-o~\cite{chen2025blip3} incorporates around 10k GPT-4o~\cite{gpt4o}-generated image-text pairs during fine-tuning, leading to notable gains in instruction-following. ShareGPT-4o-Image~\cite{chen2025sharegpt} empowers Janus-Pro~\cite{chen2025janus} with advanced text-to-image and image editing capabilities by leveraging synthetic data generated by GPT-4o~\cite{gpt4o}.

However, unlike typical LLM reasoning tasks~\cite{hendrycks2021measuring,cobbe2021training,rein2024gpqa} or VQA tasks~\cite{Agrawal2015VQAVQ, ye2024loki, liu2023mmbench, zhang2024mathverse, jiang2025mme}, which often lack existing high-quality textual data, image generation tasks naturally benefit from an abundant source of high-quality data—real-world images.
Real-world images already represent the highest-quality outputs of visual synthesis, containing complex low-level details such as texture, lighting, and structural consistency. In LLM tasks, stronger models can generate higher-quality textual supervision. In contrast, for image synthesis tasks, despite continuous improvements, even the most advanced models still fall short compared to real-world images. Therefore, simply using synthetic data—produced by advanced models and similar to natural images—holds limited significance. Therefore, a natural question arises: \textbf{Why is it necessary to train on GPT-4o-generated synthetic data?}

\begin{figure}[h]
    \centering
    \includegraphics[width=1\linewidth]{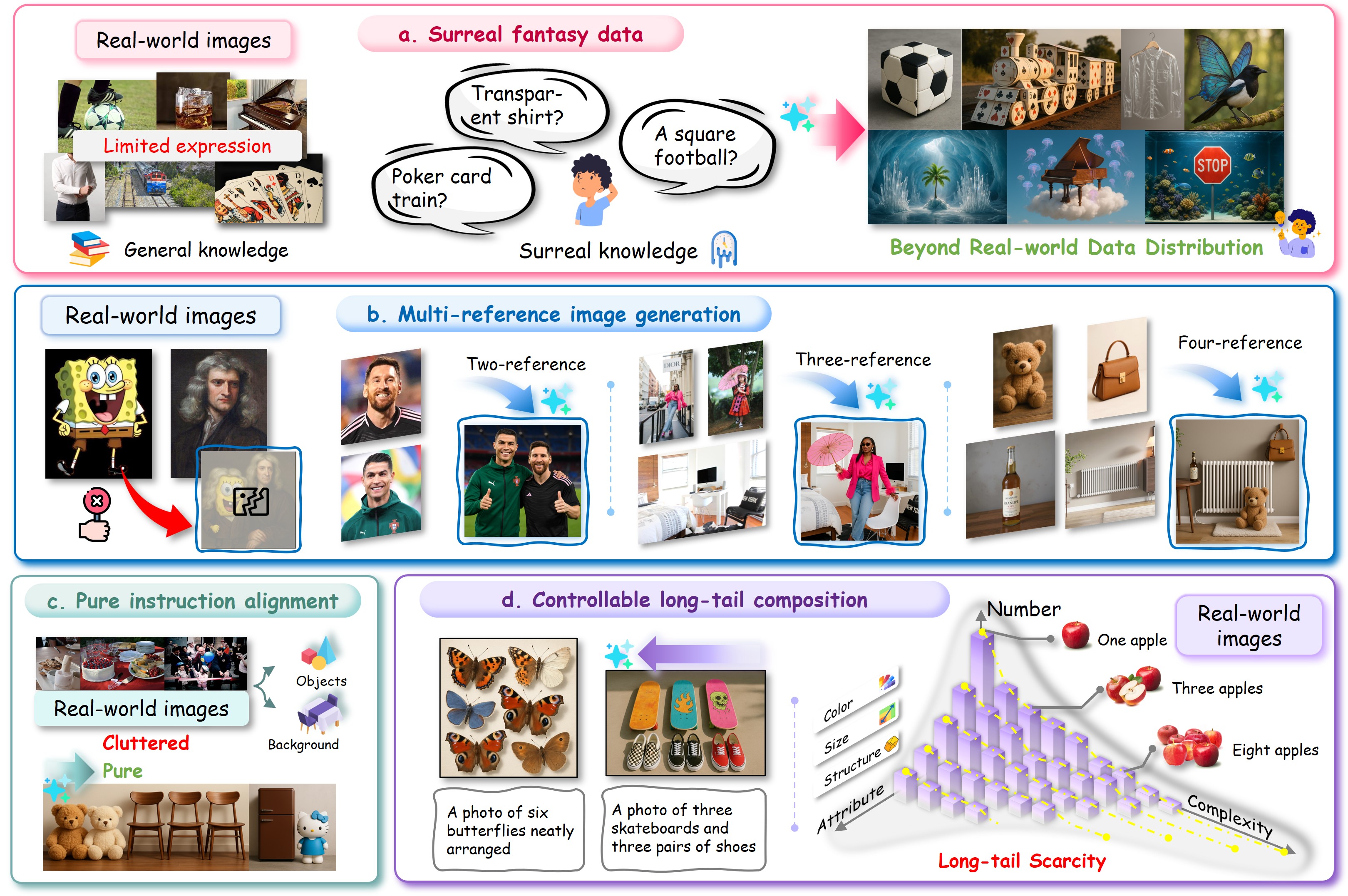}
    \vspace{-10pt}
    \caption{Illustration of the key advantages of synthetic images. architectures.}
    \label{fig:2}
    \vspace{-10pt}
\end{figure}

In this work, we identify and analyze two key advantages of synthetic images.
First, they can complement rare scenarios that are underrepresented in real-world datasets, such as surreal fantasy scenes or multi-reference image generation.
Second, synthetic images provide clear and controllable supervision. Synthetic images can provide data with pure backgrounds and long-tailed attribute combinations, thereby overcoming text–image misalignment and long-tail distribution issues in real-world images.
As illustrated in Figure~\ref{fig:2}, we elaborate on the advantages as follows:

\textbf{(a) Surreal fantasy data}. Fantastical or physically implausible content—such as a train made of playing cards—is common in creative user prompts but absent from real-world image datasets.
In contrast, advanced generative models—with their strong understanding-for-generation capabilities—can synthesize such fantastical content effectively.
Weaker models can benefit from this type of synthetic data to directly learn imaginative generation beyond the constraints of real-world distributions.

\textbf{(b) Multi-reference Generation:} Structured “Multi-to-one” image generation tasks are also lacking in natural datasets. In contrast, synthetic data allow for explicitly designed multi-input generation settings, offering greater diversity in instructions and richer reference information compared to typical data collected from video frames \cite{wu2025omnigen2}.

\textbf{(c) Pure instruction alignment:} Real-world images often contain cluttered backgrounds or irrelevant objects, while the accompanying text omits these details, causing vision–language misalignment and making it harder for models to learn accurate instruction following. In contrast, synthetic images provide clean, controlled supervision—typically showing only the target object against a simple background. For example, a cabinet and a Hello Kitty doll placed in front of a plain wall create a focused and unambiguous training instance, reducing alignment gap.

\textbf{(d) Controllable long-tail composition:} Synthetic data allows precise control over attributes such as color, quantity, and spatial arrangement, enabling coverage of complex long-tail combinations. For example, “eight red apples” represents a long-tail case compared to more frequent quantities like "one" or "two" apples. Despite its scarcity in the real world, such scenarios can be easily synthesized. This control enhances a model's ability to follow complex and corner-case instructions.

Building upon these insights, we present \textbf{Echo-4o-Image}, a new synthetic dataset generated using the advanced closed-source model GPT-4o.
By harnessing the power of synthetic image data via the echoes of GPT-4o, this dataset aims to enhance current image generation models by addressing the inherent limitations of real-world image datasets.
Echo-4o-Image includes 38K surreal fantasy samples, 73K multi-reference image generation samples, and 68K complex instruction-following samples, collectively enriching model capabilities in imagination, mulit-reference, and instruction alignment. Based on this dataset, we fine-tune the unified multimodal generation baseline Bagel~\cite{bagel} to develop a new model, \textbf{Echo-4o}, which exhibits remarkable improvements on both text-to-image generation and multi-reference image synthesis tasks. 

To rigorously evaluate the model’s instruction-following and imaginative generation, we further introduce two novel benchmarks: \textit{GenEval++} and \textit{Imagine-Bench}. GenEval++ incorporates an automated evaluator powered by GPT-4.1 and significantly increases the difficulty and compositional complexity of test instructions, addressing the limitations of scoring saturation and insufficient accuracy found in existing text-to-image evaluations. Imagine-Bench focuses on imaginative generation, offering a comprehensive evaluation of conceptual creativity and visual consistency across three dimensions: fantasy fulfillment, identity preservation, and aesthetic quality.

We conduct extensive experiments on Echo-4o to demonstrate the effectiveness of Echo-4o-Image.
As shown in Figure~\ref{fig:teaser}(a), Echo-4o achieves notable gains on mainstream benchmarks, especially in following complex instructions and generating high-quality imaginative images. It also unlocks multi-reference generation, excelling in multi-image fusion and visual consistency. Figure~\ref{fig:teaser}(b) further shows that applying the Echo-4o-Image dataset to other foundation models (e.g., OmniGen2, BLIP3-o) consistently improves performance across evaluation dimensions, highlighting the dataset’s generalizability and transferability. Our main contributions are summarized as follows:

\begin{itemize}
  \item We analyze and summarize the advantages of synthetic data over real-world images, highlighting its ability to generate rare scenarios and to provide pure, long-tailed supervision for instruction-following tasks.

  \item We curate \textit{Echo-4o-Image}, a synthetic dataset of ~180K samples generated using GPT-4o, covering surreal scenes, multi-reference generation, and instruction-following tasks.
        
  \item We fine-tune the Bagel model on \textit{Echo-4o-Image}, yielding the unified generative model \textit{Echo-4o}, which achieves state-of-the-art performance across multiple benchmarks. Furthermore, \textit{Echo-4o-Image} consistently enhances other backbone models such as OmniGen2 and BLIP3-o, demonstrating strong transferability.

  \item We propose two new evaluation benchmarks: \textit{GenEval++} increases instruction complexity to alleviate score saturation in text-to-image evaluation. \textit{Imagine-Bench} targets fantasy tasks and is designed to assess both understanding and generation of imaginative content.
\end{itemize}

\section{Echo-4o-Imgae}

We first introduce Echo-4o-Image, a large-scale synthetic dataset distilled from GPT-4o. As illustrated in Figure~\ref{fig:dataset}, the dataset contains approximately 179,000 samples spanning three distinct task types: 38K surreal fantasy generation tasks, 73K multi-reference image generation tasks, and 68K complex instruction execution tasks. Notably, the surreal fantasy and multi-reference generation subsets consist of rare and underrepresented data in existing resources. 
In the following sections, we provide a detailed illustration of the dataset construction pipeline and strategies.

\subsection{Surreal Fantasy Image Generation}

We curate a specialized subset of text-to-image data focused on surreal fantasy content. The prompts in this subset involve irregular modification of the conventional attributes, times, or spaces of an object. While this data type represents a substantial portion of user requests, it is seldom encountered in real-world training data.

As shown in Figure~\ref{fig:dataset}(a), we design a structured pipeline for constructing the fantasy generation subset of Echo-4o-Image.
We begin by collecting a set of common object concepts from the COCO and Open Images datasets, which serve as the primary subjects for subsequent generation. Given an object, GPT-4o first performs identity attribute construction, outlining its canonical properties such as color, shape, or size.
It then conducts conceptual deformation, creatively altering and recombining these attributes to introduce novel and imaginative features.
These deformations fall into three primary categories:

\textbf{Attribute shift} - altering conventional object features, such as color, shape, or size (e.g., a white banana, a cube-shaped soccer ball, or a palm-sized giraffe).

\textbf{Hybridization} – redefining the material of an object (e.g., a tomato made of crystal) or combining disparate objects (e.g., a house made of bananas).

\textbf{Spatiotemporal anomaly} – placing objects in impossible settings (e.g., a train in the clouds) or merging eras (e.g., ancient artifacts with futuristic technology).

In addition to single-object prompts, we further extend to \textbf{Multi-object} fantasy compositions, in which GPT-4o generates surreal instructions involving interactions between multiple entities. The resulting imaginative prompts are then passed to GPT-4o for final image generation, yielding visually coherent and creatively rich samples.

\subsection{Multi-Reference Image Generation}

Multi-reference image generation takes several reference images as input, along with a text prompt. The prompt specifies which elements to extract from each image and how to combine them into a coherent output. This task requires strong understanding capabilities to interpret complex prompts. It also demands exceptional generation abilities to preserve distinctive features from each reference. These features must be seamlessly integrated into the output image. Like surreal fantasy generation, limited ready-to-use training data exists for this challenging task.

\begin{figure}
    \centering
    \includegraphics[width=1\linewidth]{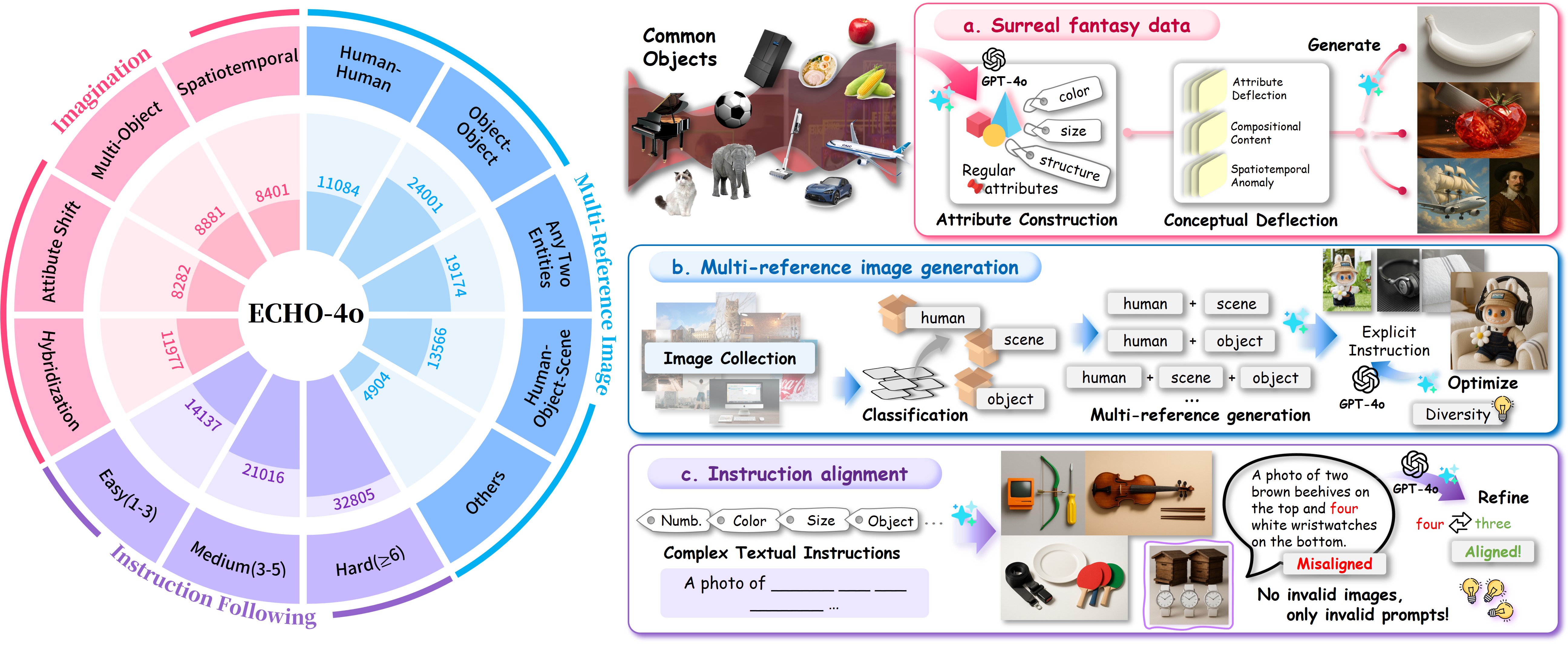}
    
    \vspace{-2pt}
    \caption{Overview and construction pipeline of the Echo-4o-Image dataset.}
    \label{fig:dataset}
    \vspace{-10pt}
\end{figure}

As shown in Figure~\ref{fig:dataset}(b), for the Multi-reference image generation task, we design reference combinations involving people, objects, and scenes, with each sample containing two to four input images.
We first curate a diverse collection of reference images covering a wide range of categories, including portraits, street photography, animals, objects, clothing and accessories, natural landscapes, famous landmarks, and indoor scenes.
This diversity in input references provides a strong foundation for generating outputs that are both creatively rich and highly varied.

Using GPT-4o, we generate instructions conditioned on these reference images. Each instruction is tailored to specific interaction types (e.g., Human–Object, Object–Scene), and explicitly references the image indices (e.g., \textit{Image\_1}, \textit{Image\_2}) to reduce ambiguity and improve alignment between the instruction and the visual inputs. After generating the target image with GPT-4o, we further refine the original instruction through rewriting strategies aimed at improving linguistic diversity and semantic clarity. During this process, explicit references such as \textit{Image\_}1 or \textit{Image\_2} may be replaced with specific descriptions of the corresponding people or objects depicted. This enhances both the quality of training data and the model’s generalization ability in multi-image generation tasks.

\subsection{Instruction-Following Data Generation}

For the text-to-image instruction-following task, we synthesize data by introducing more complex multi-object and multi-attribute instructions.
As shown in Figure~\ref{fig:dataset}(c), we begin with a carefully curated set of base object concepts and adopt a template-driven generation strategy to systematically construct prompts involving attributes such as color, position, count, and size, thereby producing semantically rich and diverse instructions. Subsequently, we employ GPT-4o to generate images. We summarize two key advantages of our synthesized data over real-world data:

\paragraph{Pure instruction alignment.}  Compared to real-world images, synthetic data generated by GPT-4o typically features cleaner backgrounds and the absence of irrelevant objects. For example, when generating an image of a violin alongside two pairs of chopsticks, these objects are rendered directly against a neat and uncluttered background. Remarkably, GPT-4o can achieve this even without any additional prompt constraints. Such visually clean images facilitate clearer conceptual mappings between the objects described in the prompt and their corresponding visual representations, thereby reducing the learning difficulty in instruction-following tasks.

\paragraph{Controllable long-tail composition.} 
Compared to Geneval, which primarily contains a limited set of semantic concepts, our dataset substantially increases instruction complexity.
For instance, one of the most challenging prompts in GenEval may contain only four semantic units—e.g., “an orange television and a green bow”.
In contrast, our dataset includes prompts such as “an orange television, a green bow, and a yellow screwdriver” or “a belt, a plate, and three table tennis paddles”. The increase in both the number of objects and their attributes results in more intricate attribute combinations, thereby addressing the scarcity of such long-tailed data in real-world image distributions.

Although we employ the state-of-the-art GPT-4o model for image generation, alignment errors may still occur—i.e., the generated image may not fully match the original prompt.
To address this issue, we introduce a text rewriting strategy to ensure data usability.
The core principle is: “There are no invalid images, only invalid text.”
When a misalignment is detected, we revise the original text based on the generated image to ensure that each image–instruction pair forms a semantically valid and consistent training sample.
For example, in Figure 4, if the generated image contains three watches but the text specifies four, the description is corrected to three.
Rather than discarding misaligned samples, we rewrite the text, allowing us to fully leverage the valuable GPT-4o synthetic data.

\section{Echo-4o}
To validate the effectiveness of our curated Echo-4o-Image dataset, we fine-tune Bagel~\cite{bagel}, a strong baseline model, to obtain Echo-4o—a unified multimodal generative model that excels in both text-to-image and multi-reference generation tasks.

Bagel is a unified multimodal generative model capable of both image understanding and generation. The model supports text-to-image generation and single image-to-image generation tasks, including image editing and free-form manipulation. Architecturally, Bagel employs a ViT~\cite{dosovitskiy2020image, zhai2023sigmoid} for image understanding and a VAE~\cite{kingma2013auto} for image generation, utilizing a mixture of transformers approach where one expert processes VAE tokens while another handles all other tokens. For multi-reference tasks, both ViT and VAE feature of the image is input to the model. However, although multi-image input is architecturally feasible, Bagel demonstrates poor performance on multi-reference generation tasks.

We fine-tune Bagel using all the text-to-image and multi-reference data from Echo-4o-Image. The training objective uses flow matching loss calculated exclusively on the output image. We train all the model components except the VAE for 24,000 steps with a learning rate of 2e-5. Through this fine-tuning process, Echo-4o achieves exceptional performance on multi-reference generation while further enhancing text-to-image capabilities, as detailed in Section~\ref{sec:experiment}.
We deliberately select Bagel as our baseline due to its training on trillions of tokens of interleaved multimodal data. The fact that Echo-4o-Image still yields significant improvements over this extensively trained model demonstrates the complementary value of carefully designed synthetic data.
\section{GenEval++ \& Imagine-Bench}

\subsection{Instruction-Following Evaluation — GenEval++}

\begin{figure}[h]
    \centering
    \includegraphics[width=0.95\linewidth]{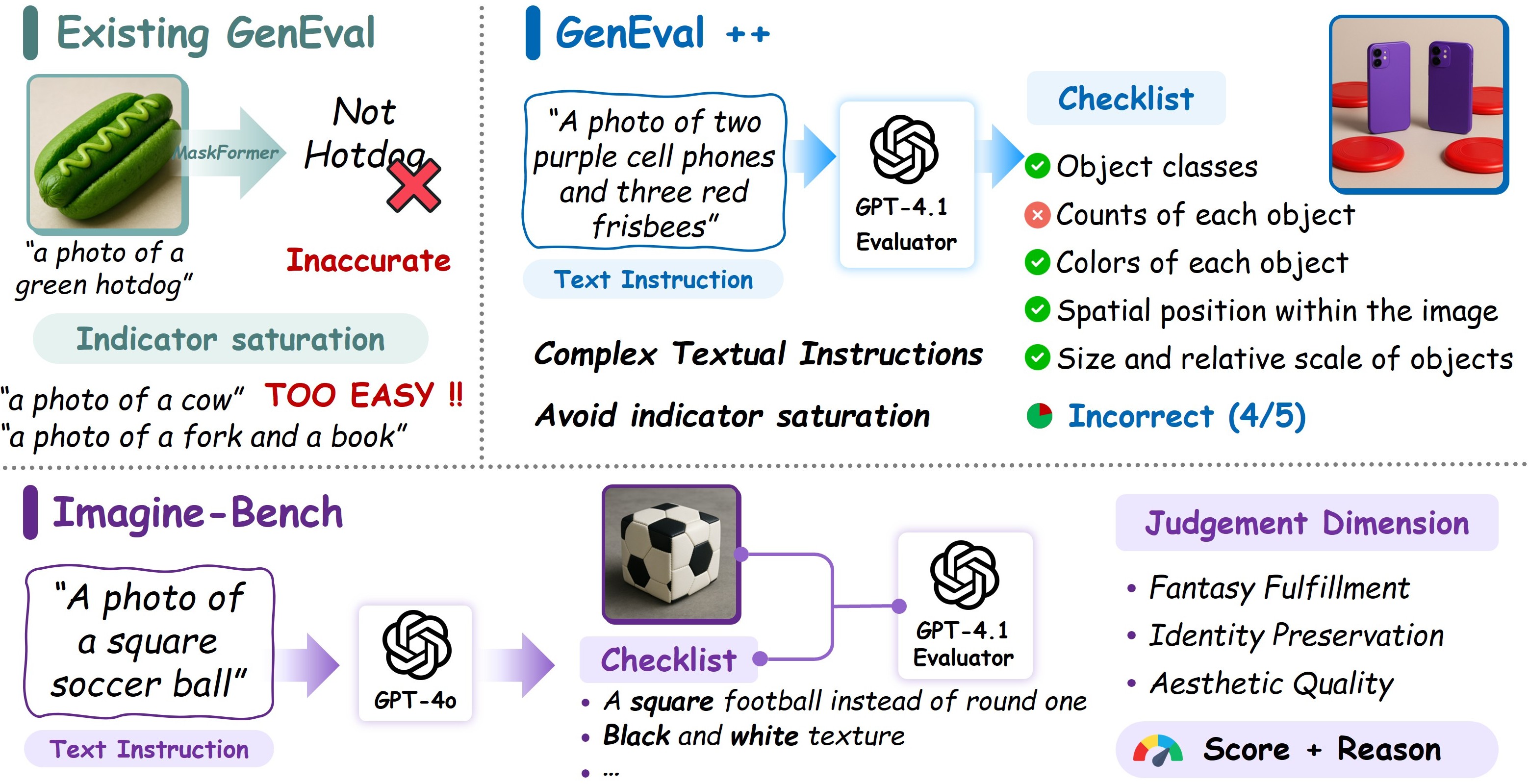}
    \vspace{-5pt}
    \caption{Overview of the Proposed Benchmarks: GenEval++ and Imagine-Bench.}
    \label{fig:Benchmark}
    \vspace{-10pt}
\end{figure}

Previous instruction-following benchmarks, such as GenEval~\cite{ghosh2023geneval}, have been widely adopted to evaluate the capability of image generation models to follow textual instructions.
However, these benchmarks typically rely on object detectors or CLIP-based models for automatic scoring, both of which exhibit substantial limitations in accuracy.
As illustrated in Figure \ref{fig:Benchmark}, within GenEval, when evaluating prompts such as “green hotdog”, the detector frequently produces incorrect judgments—despite the generated image being visually correct—due to the weak association between “hotdog” (a type of food) and the color green.
Similarly, occlusions between objects can hinder accurate counting, resulting in false negatives.
Moreover, the textual instructions in existing benchmarks are relatively simple and involve limited semantic diversity.
Consequently, current models often achieve scores in the range of 0.8–0.9, indicating metric saturation and thus constraining the discriminative power of these benchmarks.

To address these limitations, \textbf{we introduce GenEval++, a more accurate and challenging benchmark for evaluating instruction fidelity in image generation.}
As illustrated in Figure~\ref{fig:Benchmark}, GenEval++ employs the GPT-4.1~\cite{gpt4-1} multimodal model as the evaluator, leveraging its strong capability to understand over complex semantic compositions to assess the consistency between generated images and textual instructions.
Following a predefined checklist covering multiple criteria—Object, Counts, Color, Position, and Size—the evaluator only marks a result as correct if all conditions are satisfied.
Furthermore, the benchmark covers seven task types involving different attribute combinations, each comprising 40 high-complexity prompts, for a total of 280 textual instructions. GenEval++ features richer semantics and more diverse compositions, resulting in a task difficulty that is significantly higher than that of the original GenEval.
Additionally, to align with the prompt style "A photo of", outputs rendered in anime style or containing multiple disjoint elements are regarded as invalid.

\subsection{Surreal \& Fantasy Evaluation — Imagine-Bench}

Moreover, existing evaluation protocols predominantly focus on real-world generation tasks.
However, the true value of generative models lies not only in reproducing reality but also in creating the unknown, which aligns with a substantial portion of user-driven creative instructions.
To this end, we introduce a new benchmark, \textbf{Imagine-Bench, designed to evaluate a model’s capability in surreal and imaginative image generation.}
The primary tasks in Imagine-Bench involve augmenting common objects with fantastical elements while preserving their core identity features.
For example, the instruction “a square soccer ball” requires the model to alter the shape to a cube while retaining the standard black-and-white soccer texture.
Such tasks present a rigorous challenge for current understanding–generation unified models, as they demand breaking entrenched associations between concepts and appearances to enable genuine creative synthesis.

Imagine-Bench comprises 270 diverse creative instructions encompassing a wide range of surreal attributes. During evaluation, each instruction is first processed by GPT-4o to generate a corresponding checklist containing detailed explanations and expected outcomes, including the required fantastical modifications and a description of the object’s invariant identity features. Given the prompt and the generated image, GPT-4.1 assigns scores along three dimensions:
\textit{(1) Fantasy Fulfillment} – whether the generated image faithfully realizes the surreal aspects of the prompt;
\textit{(2) Identity Preservation} – whether the transformed object retains the essential visual characteristics of its original identity;
\textit{(3) Aesthetic Quality} – the visual appeal, creativity, and diversity of the generated image.

Inspired by VIEScore~\cite{ku2023viescore} and the evaluation protocol of OmniGen2, GPT-4.1 scores each dimension on 0–10 and provides explicit reasoning for each score, ensuring both rigor and interpretability in the evaluation. We further adopt a more stringent scoring scheme, where the final score is computed as 0.8 × min(Fantasy Fulfillment, Identity Preservation) + 0.2 × Aesthetic Quality.

\section{Experiments}
\label{sec:experiment}

In this section, we present a comprehensive evaluation of Echo-4o, focusing on its performance across a diverse set of generation tasks, including instruction-following image generation, surreal/fantasy image synthesis, and multi-reference image generation. The results demonstrate that Echo-4o consistently achieves strong performance across these tasks, highlighting the effectiveness of the Echo-4o-Image synthetic dataset in enhancing generative model capabilities.

\subsection{Instruction-Following Image Generation}
\label{subsec:t2i}

We evaluate the instruction-following capability of our model on two widely used benchmarks—GenEval and DPG-Bench, as well as our newly proposed benchmark, GenEval++. As shown in Table~\ref{tab:geneval}, Echo-4o achieves a score of 0.89 on GenEval, outperforming prior state-of-the-art unified models such as Bagel and OmniGen2. On DPG-Bench (Table~\ref{tab:dpgbench}), Echo-4o attains an overall score of 86.07, outperforming strong competitors including SD3 and UniWorld. These results demonstrate that Echo-4o delivers consistently superior performance across benchmarks of different types, indicating its strong instruction-following generation capabilities or both short-text and complex long-text instructions.

In existing text-to-image instruction-following tasks, Echo-4o achieves consistently superior performance.
Compared to the baseline model Bagel, Echo-4o delivers an 8.5\% improvement on GenEval, validating the effectiveness of the pure synthetic data in the Echo-4o-Image dataset for enhancing instruction-following capabilities.
These clean-background samples enhance the model’s ability to learn precise alignments between images and textual instructions.
Although the training data is predominantly composed of short-text instructions, the model also demonstrates strong generalization to complex long-text generation tasks, as evidenced by its performance on DPG-Bench.

\begin{table}[!t]

    \centering
    \resizebox{0.99\linewidth}{!}{
        \begin{tabular}{lcccccccc}
            \toprule
            Method & \multicolumn{1}{c}{Single object} & \multicolumn{1}{c}{Two object} & \multicolumn{1}{c}{Counting} & \multicolumn{1}{c}{Colors} & \multicolumn{1}{c}{Position} & \multicolumn{1}{c}{Color attribution} & \multicolumn{1}{c}{Overall} \\
            \midrule
            SDv2.1~\cite{rombach2022high} & 0.98 & 0.5 & 0.44 & 0.85 & 0.07 & 0.17 & 0.50 \\
            SDXL~\cite{podell2023sdxl} & 0.98 & 0.74 & 0.39 & 0.85 & 0.15 & 0.23 & 0.55 \\
            IF-XL & 0.97 & 0.74 & 0.66 & 0.81 & 0.13 & 0.35 & 0.61 \\
            LUMINA-Next~\cite{zhuo2024lumina} & 0.92 & 0.46 & 0.48 & 0.70 & 0.09 & 0.13 & 0.46 \\
            SD3-medium~\cite{sd3-medium} & 0.99 & 0.94 & 0.72 & 0.89 & 0.33 & 0.60 & 0.74 \\
            FLUX.1-dev~\cite{FLUX} & 0.99 & 0.81 & 0.79 & 0.74 & 0.20 & 0.47 & 0.67 \\
            OmniGen~\cite{xiao2025omnigen} & 0.98 & 0.84 & 0.66 & 0.74 & 0.40 & 0.43 & 0.68 \\ 
            \midrule
            TokenFlow-XL~\cite{qu2025tokenflow} & 0.95 & 0.60 & 0.41 & 0.81 & 0.16 & 0.24 & 0.55 \\ 
            Janus~\cite{wu2025janus} & 0.97 & 0.68 & 0.30 & 0.84 & 0.46 & 0.42 & 0.61 \\
            Janus Pro~\cite{chen2025janus} & 0.99 & 0.89 & 0.59 & 0.90 & 0.79 & 0.66 & 0.80 \\
            Emu3-Gen~\cite{wang2024emu3} & 0.98 & 0.71 & 0.34 & 0.81 & 0.17 & 0.21 & 0.54 \\
            Show-o~\cite{xie2024show} & 0.98 & 0.80 & 0.66 & 0.84 & 0.31 & 0.50 & 0.68 \\
            MetaQuery-XL~\cite{pan2025metaquery} & - & - & - & - & - & - & 0.80 \\
            BLIP3-o 8B~\cite{chen2025blip3} & - & - & - & - & - & - & 0.84 \\
            UniWorld-V1~\cite{lin2025uniworld} & 0.99 & 0.93 & 0.79 & 0.89 & 0.49 & 0.70 & 0.80 \\
            OmniGen2~\cite{wu2025omnigen2} & \textbf{1} & 0.95 & 0.64 & 0.88 & 0.55 & 0.76 & 0.80 \\
            GPT-4o-Image & 0.99 & 0.92 & 0.85 &  0.92 &  0.75 &  0.61 & 0.84 \\
            BAGEL~\cite{bagel} & 0.99 & 0.94 & 0.81 & 0.88 & 0.64 & 0.63 & 0.82 \\
            \rowcolor{myblue}\textbf{Echo-4o} & \textbf{1} & \textbf{0.97} & \textbf{0.83} & \textbf{0.95} & \textbf{0.86} & \textbf{0.77} & \textbf{0.89} \\         
            \bottomrule
        \end{tabular}
    }
    \vspace{5pt}
    \caption{Evaluation of text-to-image generation ability on GenEval~\cite{ghosh2024geneval} benchmark. Results of GPT-4o-Image are tested by~\cite{yan2025gpt}.}
    \label{tab:geneval}
    	\vspace{-15pt}
\end{table}

\begin{table}[t]
    \centering
    \begin{tabular}{lcccccc}
        \toprule
        Method & \multicolumn{1}{c}{Global$\uparrow$} & \multicolumn{1}{c}{Entity$\uparrow$} & \multicolumn{1}{c}{Attribute$\uparrow$} & \multicolumn{1}{c}{Relation$\uparrow$} & \multicolumn{1}{c}{Other$\uparrow$} & \multicolumn{1}{c}{Overall$\uparrow$} \\
        \midrule
        SDXL~\cite{podell2023sdxl} & 83.27 & 82.43 & 80.91 & 86.76 & 80.41 & 74.65 \\ 
        Hunyuan-DiT~\cite{li2024hunyuan} & 84.59 & 80.59 & 88.01 & 74.36 & 86.41 & 78.87 \\
        DALLE3~\cite{dalle3} & 90.97 & 89.61 & 88.39 & 90.58 & 89.83 & 83.50 \\
        SD3-medium~\cite{sd3-medium} & 87.90 & 91.01 & 88.83 & 80.70 & 88.68 & 84.08 \\
        FLUX.1-dev~\cite{FLUX} & 82.1 & 89.5 & 88.7 & 91.1 & 89.4 & 84.0 \\ 
        OmniGen~\cite{xiao2025omnigen} & 87.90 & 88.97 & 88.47 & 87.95 & 83.56 & 81.16 \\
        \midrule
        Show-o~\cite{xie2024show} & 79.33 & 75.44 & 78.02 & 84.45 & 60.80 & 67.27 \\
        EMU3~\cite{wang2024emu3} & 85.21 & 86.68 & 86.84 & 90.22 & 83.15 & 80.60 \\
        TokenFlow-XL~\cite{qu2025tokenflow} & 78.72 & 79.22 & 81.29 & 85.22 & 71.20 & 73.38 \\ 
        Janus Pro~\cite{chen2025janus} & 86.90 & 88.90 & 89.40 & 89.32 & 89.48 & 84.19 \\
        T2I-R1~\cite{jiang2025t2i} & 91.79 & 90.23 & 89.05 & 90.13 & 89.48 & 84.76 \\
        BLIP3-o 4B~\cite{chen2025blip3} & - & - & - & - & - & 79.36 \\
        BLIP3-o 8B~\cite{chen2025blip3} & - & - & - & - & - & 81.60 \\
        UniWorld-V1~\cite{lin2025uniworld} & 83.64 & 88.39 & 88.44 & 89.27 & 87.22 & 81.38 \\
        OmniGen2~\cite{wu2025omnigen2} & 88.81 & 88.83 & 90.18 & 89.37 & \textbf{90.27} & 83.57 \\
        BAGEL~\cite{bagel} & 88.94 & 90.37 & \textbf{91.29} & 90.82 & 88.67 & 85.07 \\
        \rowcolor{myblue}\textbf{Echo-4o} & \textbf{94.85} & \textbf{91.81} & 89.68 & \textbf{91.22} & 89.43 & \textbf{86.07} \\
        \bottomrule
    \end{tabular}
    \vspace{5pt}
    \caption{Evaluation of text-to-image generation ability on DPG-Bench~\cite{hu2024ella} benchmark.}
    \vspace{-20pt}
\label{tab:dpgbench}
\end{table}

\begin{table}[t]
    \centering{
    \renewcommand{\arraystretch}{1.3}
    \resizebox{0.99\linewidth}{!}{
        \begin{tabular}{lcccccccc}
            \toprule
            Method & \multicolumn{1}{c}{Color} & \multicolumn{1}{c}{Count} & \multicolumn{1}{c}{Color/Count} & \multicolumn{1}{c}{Color/Pos} & \multicolumn{1}{c}{Pos/Count} & \multicolumn{1}{c}{Pos/Size} & \multicolumn{1}{c}{Multi-Count} & \multicolumn{1}{c}{Overall}\\
            \midrule
            SDv2.1~\cite{rombach2022high} & 0.000 & 0.325 & 0.025 & 0.000 & 0.000 & 0.025 & 0.075 & 0.064 \\
            SDXL~\cite{podell2023sdxl} & 0.050 & 0.375 & 0.000 & 0.000 & 0.000 & 0.000 & 0.000 & 0.061 \\
            SD3-medium~\cite{sd3-medium} & 0.550 & 0.500 & 0.125 & 0.350 & 0.175 & 0.150 & 0.225 & 0.296 \\
            FLUX.1-Kontext~\cite{FLUX} & 0.425 & 0.500 & 0.200 & 0.250 & 0.300 & 0.400 & 0.325 & 0.343 \\
            FLUX.1-dev~\cite{FLUX} & 0.350 & \underline{0.625} & 0.150 & 0.275 & 0.200 & 0.375 & 0.225 & 0.314 \\
            \midrule
            GPT-4o~\cite{gpt4o} & \textbf{0.900} & \textbf{0.675} & \textbf{0.725} & \underline{0.625} & \underline{0.600} & \textbf{0.800} & \textbf{0.850} & \textbf{0.739} \\
            Janus-Pro~\cite{chen2025janus} & 0.450 & 0.300 & 0.125 & 0.300 & 0.075 & 0.350 & 0.125 & 0.246\\
            T2I-R1~\cite{jiang2025t2i} & 0.675 & 0.325 & 0.200 & 0.350 & 0.075 & 0.250 & 0.300 & 0.311 \\
            BLIP3-o 4B~\cite{chen2025blip3} & 0.125 & 0.225 & 0.100 & 0.450 & 0.125 & 0.550 & 0.225 & 0.257\\
            BLIP3-o 8B~\cite{chen2025blip3} & 0.250 & 0.250 & 0.125 & 0.600 & 0.125 & 0.575 & 0.225 & 0.307 \\
            OmniGen2~\cite{wu2025omnigen2} & 0.550 & 0.425 & 0.200 & 0.275 & 0.125 & 0.250 & 0.450 & 0.325 \\
            Bagel~\cite{bagel} & 0.325 & 0.600 & 0.250 & 0.325 & 0.250 & 0.475 & 0.375 & 0.371 \\
            \rowcolor{myblue}\textbf{Echo-4o} & \underline{0.800} & 0.575 & \underline{0.550} & \textbf{0.775} & \textbf{0.625} & \textbf{0.800} & \underline{0.625} & \underline{0.679} \\         
            \bottomrule
        \end{tabular}
    }
    \vspace{5pt}
    \caption{Evaluation of instruction following generation ability on Geneval++. \textbf{bold} indicates the best result, and \underline{underlined} denotes the second best.}
    \label{tab:Geneval++}
    	\vspace{-15pt}
    }
\end{table}

\begin{figure}[h]
    \centering
    \includegraphics[width=1\linewidth]{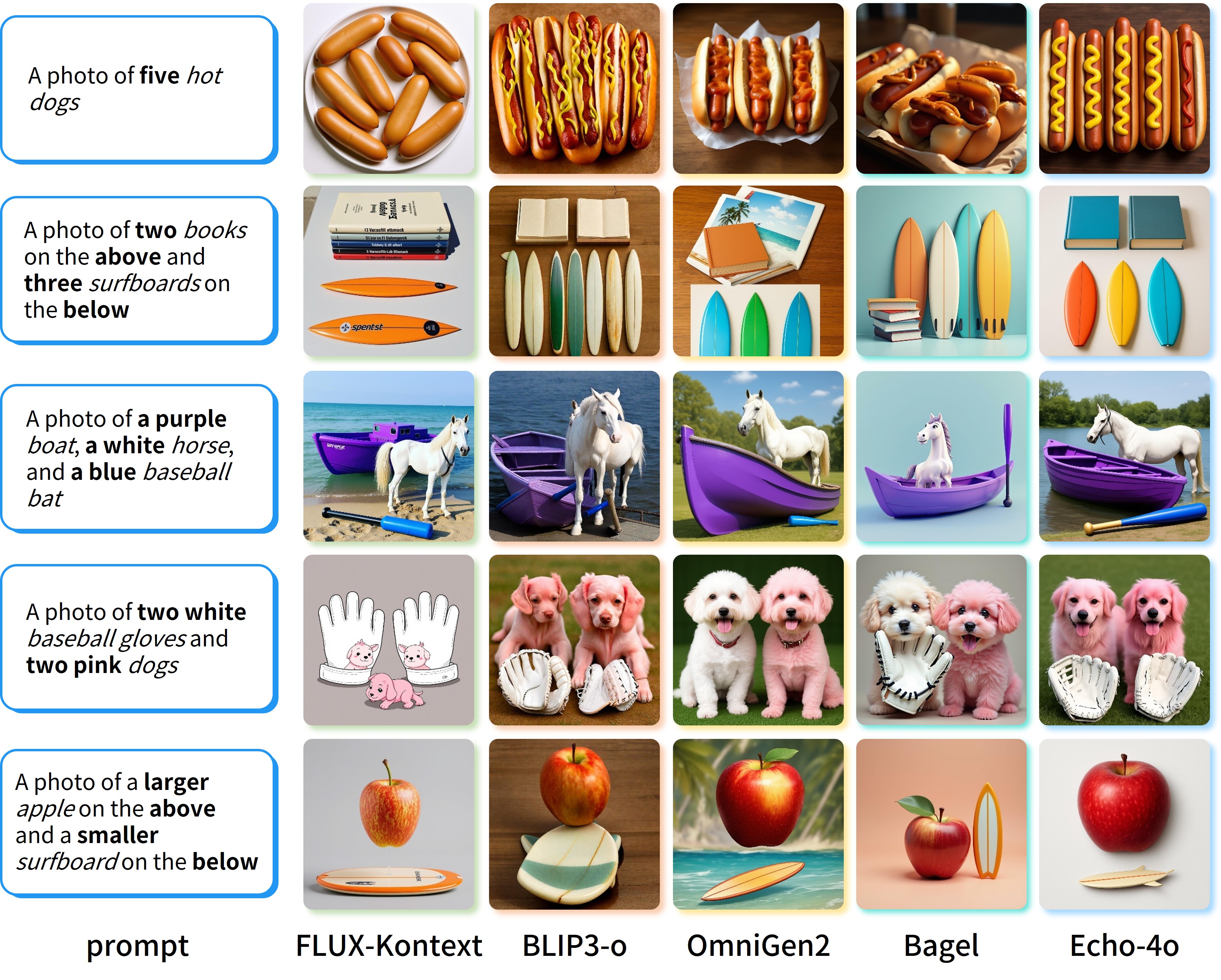}
    \vspace{-10pt}
    \caption{Qualitative comparison of different methods on GenEval++.}
    \label{fig:Geneval++}
    \vspace{-10pt}
\end{figure}

Furthermore, on the newly proposed and more challenging GenEval++ benchmark, most models perform poorly, with scores falling below 0.4.
Although GenEval++ tasks differ from those in GenEval primarily by adding only one or two additional objects and their attributes, this seemingly minor increase substantially raises task difficulty.
For example, generating five hot dogs is significantly more challenging than generating four. Early diffusion-based models, such as SDXL, almost completely fail to follow instructions in such scenarios. Even advanced unified models like Bagel and OmniGen2, which show only a small gap to GPT-4o on GenEval, fall far behind on these harder tasks.

Echo-4o achieves the best performance among all models except GPT-4o, surpassing OmniGen2 and Bagel by over 40\%.
This highlights Echo-4o’s strong instruction-following capability, which is closely related to the inclusion of more complex, long-tailed attribute data in the Echo-4o-Image.
Figure 5 further provides qualitative comparisons across different models.

\subsection{Surreal Fantasy Image Generation}

We evaluate multiple models on the Imagine-Bench benchmark to assess their understanding and creative capabilities, with the results presented in Table ~\ref{tab:Conflict-bench}.
Conventional image generation models perform poorly on this task, primarily due to their training paradigm, which often establishes a direct binding between textual concepts and visual representations.
Constrained by limited comprehension ability, these models struggle to differentiate the inherent concepts of existing objects from the additional requirements specified in fantasy-oriented instructions.
Unified models, such as BLIP3o and OmniGen2, benefit from stronger comprehension capabilities and achieve slightly better results.
Among open-source models, Echo-4o attains the best performance, directly benefiting from the inclusion of fantasy-oriented image data in Echo-4o-Image, which extends beyond the domain constraints of real-world images and thereby improves performance in a relatively straightforward manner.
Future work could explore more systematic approaches to further enhance unified models’ capabilities in both understanding and creative generation.

\begin{table}[H]
    \centering
    \resizebox{0.99\linewidth}{!}{
        \begin{tabular}{lccccc}
            \toprule
            Method & \multicolumn{1}{c}{Attribute shift} & \multicolumn{1}{c}{Spatiotemporal} & \multicolumn{1}{c}{Hybridization} & \multicolumn{1}{c}{Multi-Object} & \multicolumn{1}{c}{Overall} \\
            \midrule
            SDv2.1~\cite{rombach2022high} & 4.46 & 5.06 & 4.12 & 3.49 & 4.30 \\
            SDXL~\cite{podell2023sdxl} & 4.42 & 6.32 & 4.93 & 4.50 & 4.97 \\
            SDv3-medium~\cite{sd3-medium} &  5.14 & 5.91 & 6.30 & 6.07 & 5.78 \\
            FLUX.1-Kontext~\cite{FLUX} & 5.33 & 6.49 & 5.48 & 5.34 & 5.62 \\
            FLUX.1-dev~\cite{FLUX} & 5.68 & 7.13 & 6.38 & 5.24 & 6.06 \\
            \midrule
            GPT-4o~\cite{gpt4o} & \textbf{8.54} & \textbf{9.18} & \textbf{8.57} & \underline{7.98} & \textbf{8.56} \\
            Janus-Pro~\cite{chen2025janus} & 5.30 & 7.28 & 6.73 & 6.04 & 6.22 \\
            T2I-R1~\cite{jiang2025t2i} & 5.85 & 7.70 & 7.36 & 6.68 & 6.78\\
            BLIP3-o 4B~\cite{chen2025blip3} & 5.48 & 6.79 & 6.93 & 6.09 & 6.23 \\
            BLIP3-o 8B~\cite{chen2025blip3} & 5.80 & 7.08 & 7.06 & 6.44 & 6.51 \\
            OmniGen2~\cite{wu2025omnigen2} & 5.28 & 7.45 & 6.29 & 6.31 & 6.22 \\
            Bagel~\cite{bagel} & 5.37 & 6.93 & 6.50 & 6.41 & 6.20 \\
            \rowcolor{myblue}\textbf{Echo-4o} & \underline{7.18} & \underline{8.75} & \underline{7.52} & \textbf{8.06} & \underline{7.80} \\         
            \bottomrule
        \end{tabular}
    }
    \vspace{5pt}
    \caption{Evaluation of surreal and imaginative image generation ability on Imagine-Bench. \textbf{bold} indicates the best result, and \underline{underlined} denotes the second best.}
    \label{tab:Conflict-bench}
    \vspace{-25pt}
\end{table}

\begin{figure}[H]
    \centering
    \includegraphics[width=1\linewidth]{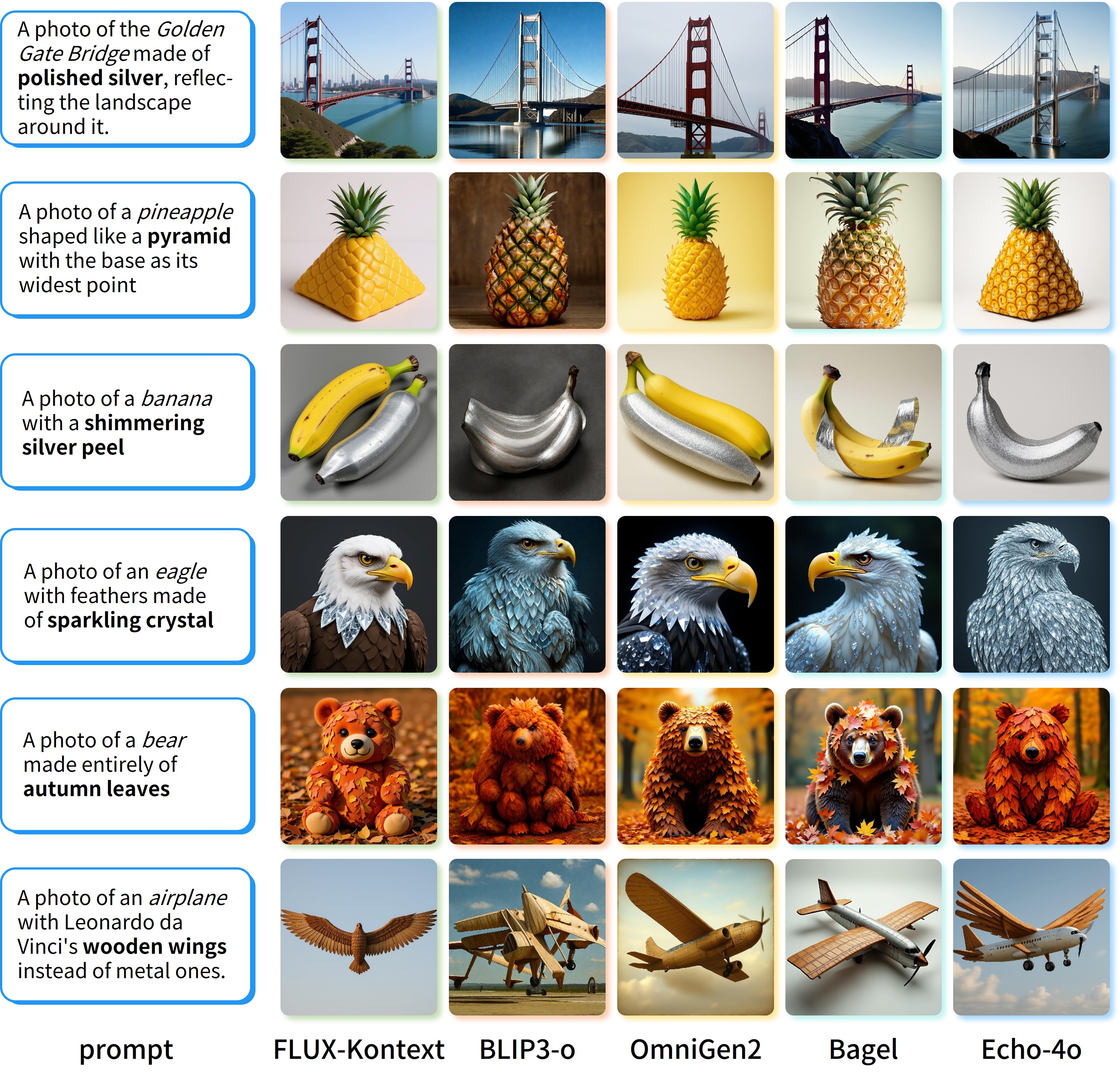}
    \vspace{-15pt}
    \caption{Qualitative comparison of different methods on Imagine-Bench.}
    \label{fig:Conflict-bench}
    \vspace{-20pt}
\end{figure}

As illustrated in Figure~\ref{fig:Conflict-bench}, some existing models either fail to respond to fantasy instructions entirely or respond only partially.
For instance, the shape of a pineapple may remain unchanged, or the Golden Gate Bridge may retain its original red color.
In contrast, Echo-4o produces results that align more closely with the intended instructions, preserving the intrinsic characteristics of the original objects while achieving higher visual aesthetics compared to models such as BLIP3o.

\subsection{Multi-Reference Image Generation}
\label{subsec:in_context}

We evaluate multi-reference image generation using the OminiContext~\cite{wu2025omnigen2} benchmark. This capability remains largely underexplored in existing image generation models and unified architectures. Among open-source models, only OmniGen2 has made initial attempts in this direction; most models, including FLUX and Bagel, either lack native support for this functionality or are entirely incompatible with multi-reference scenarios.
By leveraging synthetic data designed for multi-reference scenarios, Echo-4o obtains this capability absent in the base Bagel architecture. 

\begin{figure}[h]
    \centering
    \includegraphics[width=1\linewidth]{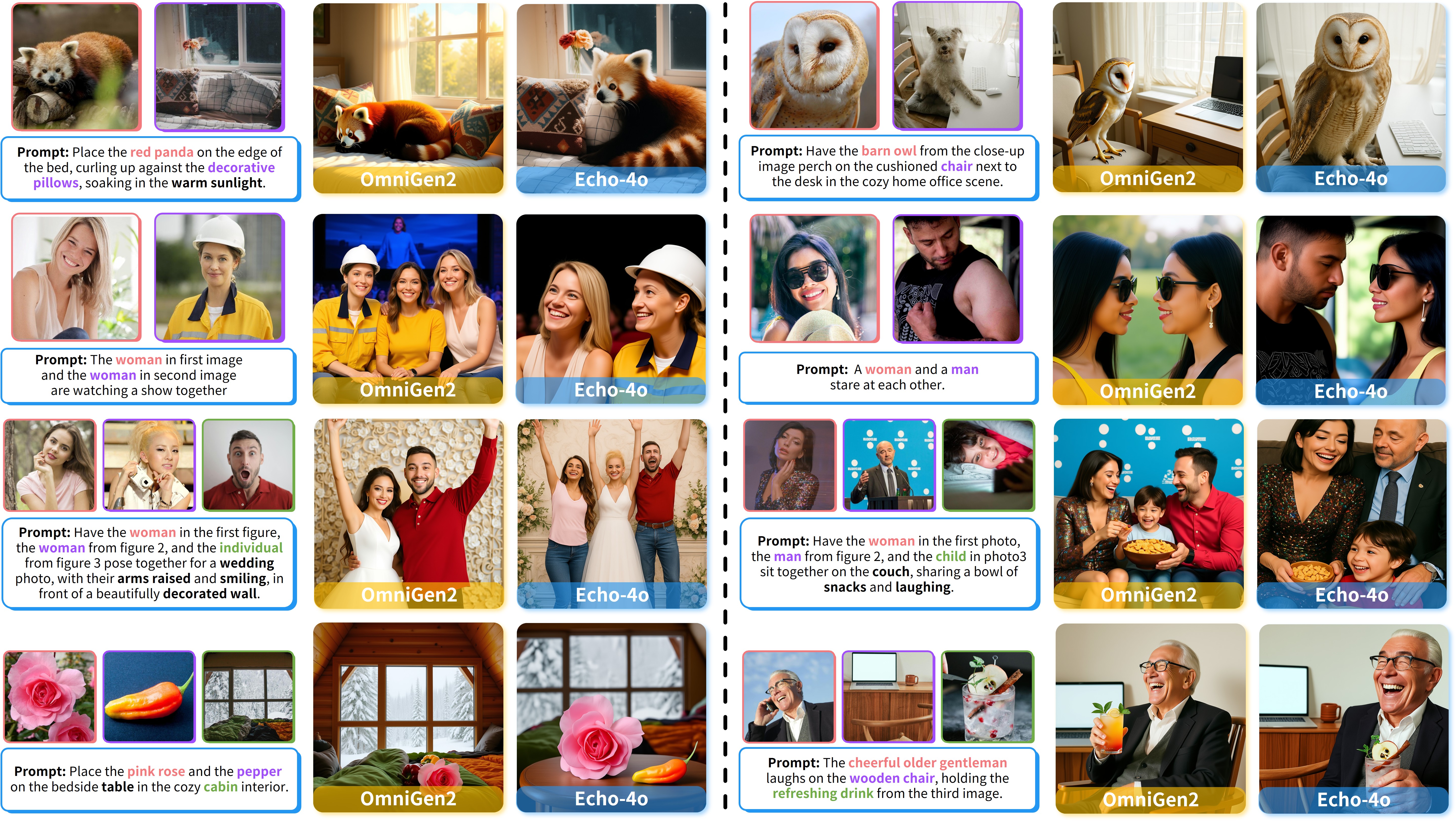}
    
    \vspace{-5pt}
    \caption{Qualitative comparison of different methods on OmniContext.}
    \label{fig:omnicontext}
    \vspace{-5pt}
\end{figure}

As shown in Table~\ref{tab:omni_context}, Echo-4o achieves the best performance among all open-source models under both MULTIPLE and SCENE settings, representing a substantial improvement over the Bagel baseline and surpassing the previous best open-source model, OmniGen2. Figure~\ref{fig:omnicontext} further provides qualitative comparisons, highlighting Echo-4o’s advantages in multi-reference generation with two or three reference images.
In terms of both instruction adherence and fidelity to the content of the reference images, Echo-4o consistently outperforms OmniGen2~\cite{wu2025omnigen2}.

\begin{table}[t]
    \renewcommand{\arraystretch}{1.3} 
    \centering
    \resizebox{0.99\linewidth}{!}{
    \begin{tabular}{l|ccc|ccc|c}
        \toprule
        \multirow{2}{*}{Method} & \multicolumn{3}{c|}{MULTIPLE} & \multicolumn{3}{c|}{SCENE} & \multirow{2}{*}{Average$\uparrow$}\\ 
        \cmidrule(lr){2-7}
        & Character & Object & Char. + Obj. & Character & Object & Char. + Obj. & \\
        \midrule
        Gemini-2.0-flash~\cite{gemini-2.0-flash} & 2.91 & 2.16 & 3.80 & 3.02 & 3.89 & 2.92 & 3.12\\
        GPT-4o~\cite{gpt4o} & \textbf{9.07} & \textbf{8.95} & \textbf{8.54} & \textbf{8.90} & \textbf{8.44} & \textbf{8.60} & \textbf{8.75} \\
        \midrule
        UNO~\cite{uno} & 2.54 & 6.51 & 4.39 & 2.06 & 4.33 & 4.37 & 4.03\\
        BAGEL~\cite{bagel} & 5.17 & 6.64 & 6.24 & 4.07 & 5.71 & 5.47 & 5.55\\
        OmniGen~\cite{xiao2025omnigen} & 5.65 & 5.44 & 4.68 & 3.59 & 4.32 & 5.12 & 4.8\\
       OmniGen2~\cite{wu2025omnigen2} & 7.11 & 7.13 & 7.45 & 6.38 & 6.71 & 7.04 & 6.97 \\
        \rowcolor{myblue}\textbf{Echo-4o} & \textbf{8.07} & \textbf{7.50} & \textbf{8.29} & \textbf{8.62} & \textbf{8.00} & \textbf{8.08} & \textbf{8.09} \\
        \bottomrule
    \end{tabular}
}
\vspace{5pt}
\caption{Overall score comparison of existing image generation models on OmniContext~\cite{wu2025omnigen2} benchmark (excluding SINGLE category). "Char. + Obj." indicates Character + Object. }
\label{tab:omni_context}
\vspace{-10pt}
\end{table}

\subsection{Effectiveness Across Base Models}

To further validate the broad effectiveness of the Echo-4o-Image dataset, we conduct additional experiments by fine-tuning several existing unified models using the synthetic data.
As shown in Figure~\ref{fig:teaser}(b), models including BLIP-3-o, Bagel, and OmniGen2 all benefit from Echo-4o-Image, exhibiting consistent improvements across benchmarks such as GenEval, GenEval++, DPG-Bench, and OminiContext.

These results demonstrate that Echo-4o-Image provides a generalizable enhancement to diverse base models, significantly improving their capabilities in instruction understanding, fantasy image synthesis, and multi-reference image generation. The consistent gains across tasks and architectures confirm the dataset's broad applicability and its important value in high-quality fine-tuning of unified multimodal generation models.

\subsection{Comparison with ShareGPT-4o-Image}

\begin{wrapfigure}{r}{0.6\textwidth}
  \vspace{-3mm}
  \centering
  \includegraphics[width=\linewidth]{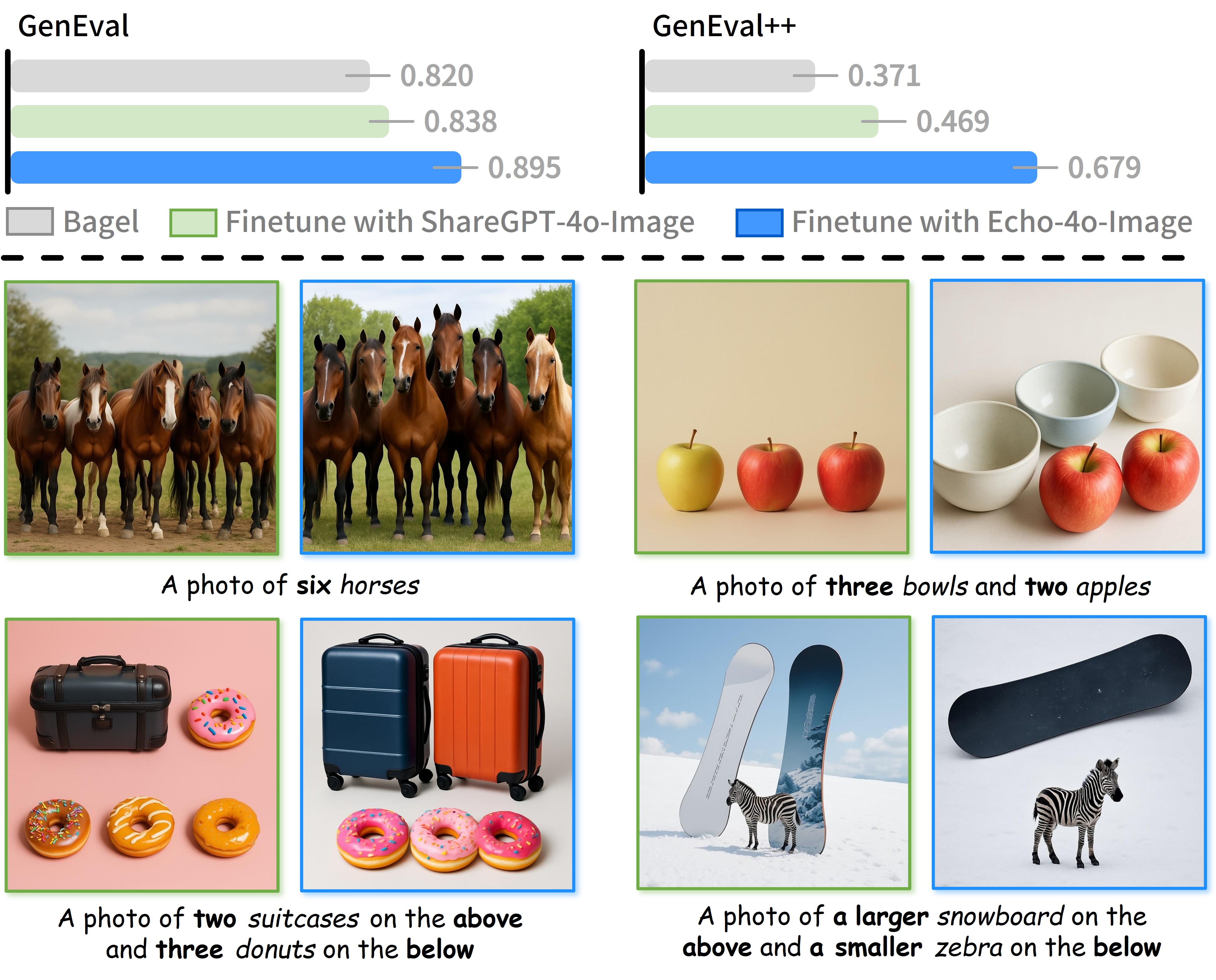 }
  \caption{
  Comparison of Echo-4o-Image and ShareGPT-4o-Image on GenEval and GenEval++.
  }
  \label{fig:sharegpt-4o-image}
  \vspace{-2mm}
\end{wrapfigure}

We further compare our dataset with ShareGPT-4o-Image, another synthetic dataset distilled from GPT-4o.
Both datasets are used to fine-tune the same baseline model, Bagel, under identical training settings until convergence.
Figure~\ref{fig:sharegpt-4o-image} reports the performance comparison on GenEval and GenEval++.

The results show that Echo-4o-Image yields substantial improvements in instruction-following capabilities, whereas ShareGPT-4o-Image exhibits only marginal gains in this aspect. After training with ShareGPT-4o-Image, the performance of Bagel on GenEval improves from 0.820 to 0.838, whereas training with Echo-4o-Image raises the score to 0.895. 
A similar trend is observed on GenEval++. Figure~\ref{fig:sharegpt-4o-image} also presents a qualitative comparison of the results.

This difference may be attributed to the fact that a considerable portion of ShareGPT-4o-Image’s data is generated from textual inputs in ALLaVA~\cite{chen2024allava}, which already contains high-quality real-world image pairs. Consequently, ShareGPT-4o-Image essentially regenerates images similar to those in real-world datasets, leading to limited additional benefits for instruction following.Nonetheless, ShareGPT-4o-Image still contributes to improving aesthetic alignment.
Notably, ShareGPT-4o-Image lacks multi-reference image generation data—a capability that Echo-4o-Image successfully enables—further underscoring its broader utility.

\section{Conclusion}

In this work, we present Echo-4o-Image, a large-scale synthetic dataset generated by GPT-4o, and demonstrate its effectiveness in enhancing unified multimodal generation models through the development of Echo-4o. Alongside, we introduce two new benchmarks, Geneval++ and Imagine-Bench, to provide a more comprehensive and challenging evaluation of image generation capabilities. The results highlight the value of the high-quality synthetic dataset Echo-4o-Image in addressing the limitations of real-world datasets and transferring effectively across different foundation models. We hope that the open-sourcing of Echo-4o-Image will advance unified multimodal generation models by complementing real-world image gaps with high-quality synthetic data, enhancing instruction following, creative generation, and multi-reference image synthesis capabilities. In future work, we plan to extend the dataset to cover image editing tasks—another scenario with limited high-quality real-world data, and to fine-tune a broader range of models, such as FLUX, to further validate its versatility and impact.

\section{Related Works}

\subsection{Image Generation Models}

Diffusion models have achieved remarkable success in the field of generative modeling, particularly in high-fidelity image synthesis~\cite{sd3,chen2023pixart,ye2024skydiffusion,brooks2023instructpix2pix,li2024crossviewdiff,controlnet,jiang2024comat,zong2024easyref,deng2024nova}. Representative systems such as the Stable Diffusion (SD) series~\cite{LDM,podell2023sdxl,sd3} and DALL·E~\cite{dalle2} have demonstrated strong text-to-image generation capabilities. More recently, research efforts have increasingly focused on multimodal generative models that aim to unify the understanding and generation of cross-modal content within a single architecture. For example, hybrid models like Show-o~\cite{xie2024show} and Transfusion~\cite{zhou2024transfusion} integrate autoregressive text generation with discrete or continuous diffusion for images in a single Transformer framework. MetaQueries~\cite{pan2025metaquery} employs learnable queries to establish an efficient interface between a frozen MLLM and a diffusion model, enabling knowledge-enhanced image generation while preserving comprehension performance. BLIP3-o~\cite{chen2025blip3} leverages a diffusion transformer to produce semantically rich CLIP image features and adopts a sequential pre-training strategy to jointly optimize understanding and generation. Models such as Bagel~\cite{bagel} and OmniGen2~\cite{wu2025omnigen2} undergo large-scale pretraining and exhibit strong generalization capabilities across diverse downstream tasks. Furthermore, several studies~\cite{guo2025cot, jiang2025t2i, tong2025delving} have introduced Chain-of-Thought (CoT) reasoning into image generation. For instance, T2I-R1~\cite{jiang2025t2i} introduces a dual-level CoT to enhance generation quality and instruction-following ability.

\subsection{Synthetic Dataset}
Leveraging synthetic data generated by larger, more capable models has emerged as a highly effective approach for enhancing the performance of weaker models. This methodology has been extensively explored across text and multimodal understanding tasks in LLMs~\cite{OpenAI2023ChatGPT, OpenAI2023GPT4TR, touvron2023llama, bai2023qwentechnicalreport} and MLLMs~\cite{openai2023gpt4v, liu2023llava, zhu2023minigpt,jiang2024mmsearch, zong2024mova}. Building upon ShareGPT, a dataset containing real conversations between users and ChatGPT, Vicuna~\cite{vicuna2023} demonstrates superior performance in generating detailed, contextually appropriate responses. For image understanding tasks, LLaVA~\cite{liu2023llava} pioneered this approach by leveraging detection dataset annotations~\cite{lin2014microsoft} to prompt ChatGPT~\cite{OpenAI2023ChatGPT} for generating detailed image captions, significantly improving multimodal understanding. This methodology was further advanced by ShareGPT-4V~\cite{chen2023sharegpt4v}, which employed GPT-4V to produce 100K high-quality image captions for training purposes. The synthetic data paradigm has since been successfully adapted across diverse domains, including mathematical problem solving~\cite{zhang2024mavismathematicalvisualinstruction}, video captioning~\cite{zhang2023video, chen2024sharegpt4video}, and 3D point cloud understanding~\cite{xu2023pointllm, guo2025pisa}.
Employing synthetic images for training has also been explored in image generation tasks. JourneyDB~\cite{sun2024journeydb} collects 4M high-quality images from various text-to-image models, demonstrating the value of synthetic visual content. More recently, the introduction of GPT-4o's image generation capabilities has opened new possibilities. ShareGPT-4o-Image~\cite{chen2025sharegpt} and GPT-Image-Edit~\cite{wang2025gpt} collect text-to-image and image editing datasets from GPT-4o. However, these works exhibit limitations in their prompt design and task formulation, potentially failing to fully leverage the powerful generative capabilities of GPT-4o for creating highly effective synthetic data.

\subsection{Image Generation Benchmark}

Early research on image generation evaluation primarily focused on assessing visual quality, with widely used metrics including FID, Inception Score (IS), and Kernel Inception Distance (KID). In recent years, driven by rapid advances in generative technology, the scope of evaluation has expanded from low-level quality metrics to measuring instruction-following capabilities. Methods such as VQAScore~\cite{lin2024evaluatingtexttovisualgenerationimagetotext}, HPSv2~\cite{wu2023humanpreferencescorev2}, and VisionReward~\cite{xu2025visionrewardfinegrainedmultidimensionalhuman} leverage learned reward models to better align with human preferences. Meanwhile, benchmarks like CompBench++\cite{huang2025t2icompbenchenhancedcomprehensivebenchmark}, GenEval\cite{ghosh2023genevalobjectfocusedframeworkevaluating}, and GenAI Bench~\cite{li2024genaibenchevaluatingimprovingcompositional} combine CLIP-based metrics with structured prompt sets to evaluate multiple compositional dimensions, such as object attributes, spatial relationships, and numerical reasoning. With the advent of MLLM-based evaluation approaches such as VIEScore~\cite{ku2023viescore}, more studies have begun to utilize large multimodal language models to assess instruction-following performance. For example, DPG-Bench~\cite{hu2024ella} incorporates complex long-form instructions and employs mPLUG-large~\cite{hu2023tifa} for evaluation; TIIF-Bench~\cite{wei2025tiif} and OmniContext~\cite{wu2025omnigen2} leverage advanced GPT-series MLLMs to achieve notable improvements in both accuracy and interpretability. However, certain existing methods (e.g., GenEval) may introduce evaluation biases due to performance limitations in their underlying detection and CLIP models. At the same time, systematic evaluation of imagination and creative generation capabilities remains underexplored, with only a few preliminary efforts, such as RF-Bench~\cite{yao2024fabrication}, addressing this direction.

\bibliography{main}
\bibliographystyle{plainnat}

\end{document}